\documentclass[10pt,twocolumn,letterpaper]{article}

\usepackage{cvpr}              

\usepackage{graphicx}
\usepackage{amsmath}
\usepackage{amssymb}
\usepackage{booktabs}
\usepackage{multirow}
\usepackage{algpseudocode}
\usepackage{algorithm}
\usepackage{bbm}
\usepackage[accsupp]{axessibility}  

\usepackage[pagebackref,breaklinks,colorlinks]{hyperref}

\usepackage[capitalize]{cleveref}
\crefname{section}{Sec.}{Secs.}
\Crefname{section}{Section}{Sections}
\Crefname{table}{Table}{Tables}
\crefname{table}{Tab.}{Tabs.}

\def\cvprPaperID{9960} 
\def\confName{CVPR}
\def\confYear{2023}

\begin{document}

\title{Probabilistic Debiasing of Scene Graphs}

\author{Bashirul Azam Biswas and Qiang Ji\\
Rensselaer Polytechnic Institute, Troy, NY-12180\\
\tt\small \{biswab, jiq\}@rpi.edu\\
}

\maketitle

\begin{abstract}
  The quality of scene graphs generated by the state-of-the-art (SOTA) models is compromised due to the long-tail nature of the relationships and their parent object pairs. Training of the scene graphs is dominated by the majority relationships of the majority pairs and, therefore, the object-conditional distributions of relationship in the minority pairs are not preserved after the training is converged. Consequently, the biased model performs well on more frequent relationships in the marginal distribution of relationships such as `on' and `wearing', and performs poorly on the less frequent relationships such as `eating' or `hanging from'.  In this work, we propose virtual evidence incorporated within-triplet Bayesian Network (BN) to preserve the object-conditional distribution of the relationship label and to eradicate the bias created by the marginal probability of the relationships. The insufficient number of relationships in the minority classes poses a significant problem in learning the within-triplet Bayesian network. We address this insufficiency by embedding-based augmentation of triplets where we borrow samples of the minority triplet classes from its neighborhood triplets in the semantic space. We perform experiments on two different datasets and achieve a significant improvement in the mean recall of the relationships. We also achieve better balance between recall and mean recall performance compared to the SOTA de-biasing techniques of scene graph models. Code is publicly available at \url{https://github.com/bashirulazam/within-triplet-debias}. 
\end{abstract}

\section{Introduction}\label{sec:intro}

\begin{figure*}[ht]
	\centering
	\includegraphics[width  =0.8\textwidth]{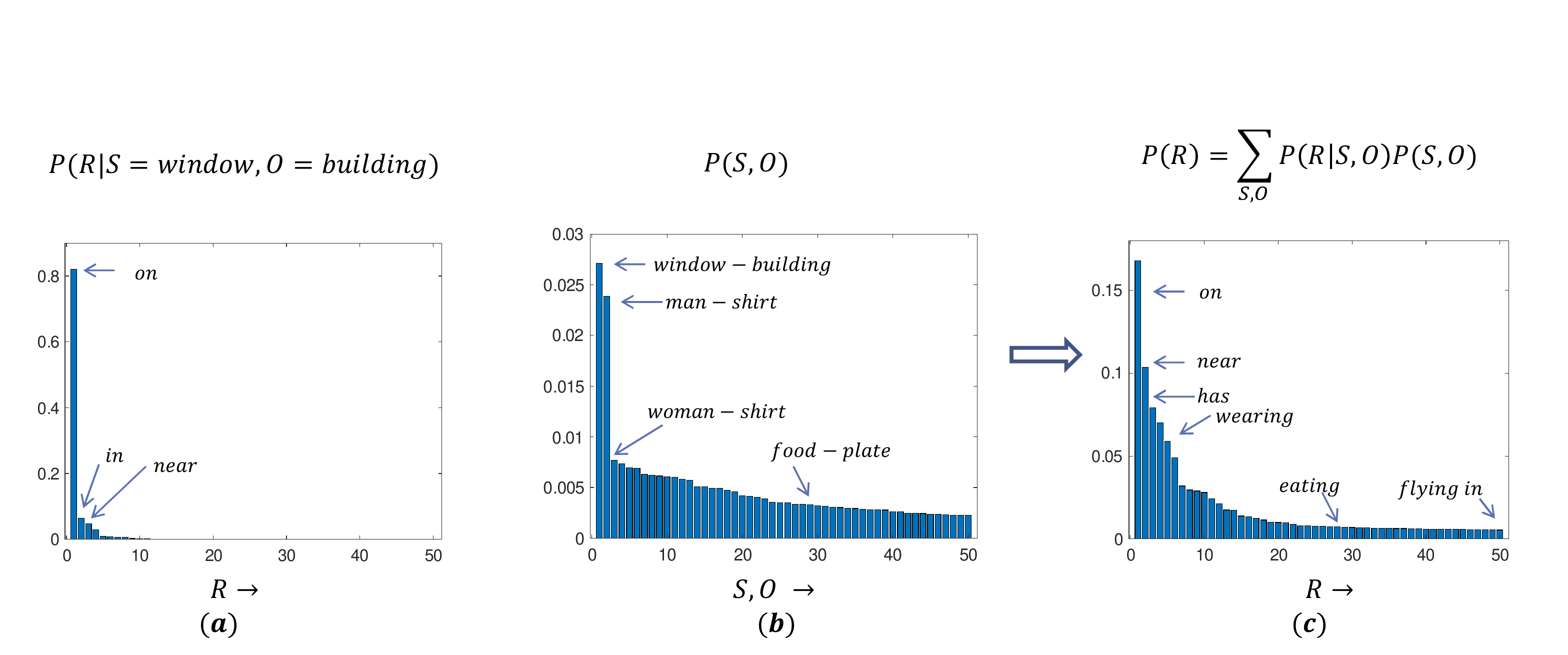}
 	\caption{(a) Within-triplet dependency of relationship on its parent object pair; (b) long-tail nature of the pair statistics where 33\% pair samples originated from top $1\%$ pairs; (c) long-tail nature of the relationships showing the dominance of `on' and `wearing'. The skewness in (c) is an effect of skewness in (b). Since `on', `near' or `wearing' dominates  in these top $1\%$ pairs, they become the majority relationships in (c) and many other relationships, such as `eating' or `flying in', which dominate in the tail pairs of (b), are suppressed in the training process.}
 	\label{figs:long_tailed}
\end{figure*}

Any visual relationship can be expressed as a triplet \textit{subject-relationship-object} and all triplets in an image can be represented as a concise graph called Scene Graph (SG) \cite{lu2016visual} where the nodes represent the objects and the edges represent relationships. This representation has been proven useful for many downstream tasks such as image captioning \cite{yang2019auto}, visual reasoning \cite{shi2019explainable}, and image generation \cite{johnson2018image}. Scene Graph Generation (SGG) has become one of the major computer vision research arenas after the introduction of Visual Genome (VG) dataset \cite{krishna2017visual}. The distribution of triplets in VG  images has two distinct characteristics: (1) the presence of strong within-triplet prior, and (2) the long-tail distribution of the relationship. As shown in Figure \ref{figs:long_tailed} (a), the within-triplet prior dictates that `window' will most likely be `on' the `building' rather than `eating' it. Zeller \textit{et al.} \cite{motif} has utilized this within-triplet prior as the conditional probability of relationships given subject and object by proposing a frequency baseline in  the SGG task. On the other hand, the distribution of relationship labels suffers from a long-tailed nature and Tang \textit{et al.} \cite{tang2020unbiased} addressed this long-tailed issue by considering a causal interpretation of the biased prediction. We argue that these two seemingly different characteristics of the relationship distribution are interrelated.  The abundance of the \textit{head} classes of the relationship distribution in Figure \ref{figs:long_tailed} (c), such as `on' and `wearing', arises from the abundance of their parent subject and object lying in the \textit{head} region of Figure \ref{figs:long_tailed} (b). 

As indicated by \cite{desai2021learning}, the long-tailed distribution exists both in relationship and object label. Since relationship labels are dependent on their object pair because of the within-triplet prior, the long-tail distribution of the relationship worsens due to the long-tail nature of the object pairs. Crowd-collection of VG images creates \textit{selection bias} and crowd-annotation of these images create \textit{label-bias} \cite{torralba2011unbiased} and \textit{co-occurring-bias} \cite{singh2020don}. To analyze such biases, we investigate the distribution of the object pair of the triplets in VG database. As shown in Figure \ref{figs:long_tailed} (b), `window-building' and `man-shirt' are the most frequently annotated pairs and top $1\%$ object pair covers $33\%$ of all triplets. As a result, the dominant relationships in these \textit{head} pairs, such as `on' and `wearing', dominate the marginal distribution of Figure \ref{figs:long_tailed} (c). 

In training a deep-learning-based SGG model, samplers will sample more relationships from the \textit{head} pairs. As a result, the Maximum Likelihood Estimation (MLE) of the parameters is biased to predict the relationship classes in the \textit{head} pairs \cite{tang2020unbiased} and the object-conditional representation of the relationship in the \textit{tail} pairs will be lost in the training process. Therefore, various deep learning-based models, which attempt to   implicitly capture such object-conditional representation  \cite{zhang2019graphical,dai2017detecting}, fail to preserve the representation in the trained model and perform poorly on the tail region of the relationships. 

Previous works attempt to retrieve the tail regions through re-sampling/re-weighting the minority classes in training \cite{desai2021learning,guo2021general,chiou2021recovering} or through causal intervention in testing \cite{tang2020unbiased}. Their success is well-demonstrated by the significant increase of minority-driven evaluation metric \textit{mean recall}. However, these approaches do not consider the strong within-triplet prior of triplets and hurt the performance of majority-driven evaluation metric \textit{recall}. Keeping this gap in mind, we propose an inference-time post-processing methodology that bolsters the minority \text{tail} classes as well as hurts the majority \text{head} classes less brutally. We propose a within-triplet Bayesian Network (BN) that combines the within-triplet prior with uncertain biased evidence from SOTA models. Posterior inference with this BN simultaneously eradicates the long-tailed bias in the marginal distribution of the relationship and restores the object-conditional within-triplet prior.

Learning such a small within-triplet BN from the training data is a seemingly trivial task where we can perform simple MLE of parameters by counting. However, because of restricting our training samples only belonging to some top-$N_r$ classes based on the marginal probability of relationship, we sacrifice many information revealing triplets in the minority pairs. For example, in the `man-pizza' pair, we see there exist many interesting relationships such as `man-biting-pizza' or `man consuming pizza' which are semantically similar to one of the top-$N_r$ valid triplets `man-eating-pizza'. This phenomenon is also a result of \textit{label bias} \cite{torralba2011unbiased} where the annotator chooses some labels over another for the same category of objects or relationships. We propose a novel method of borrowing samples from such invalid triplets into learning the distribution of the valid triplets using embedding-based augmentation.   

\begin{figure*}[ht]
	\centering
	\includegraphics[width  =0.9\linewidth]{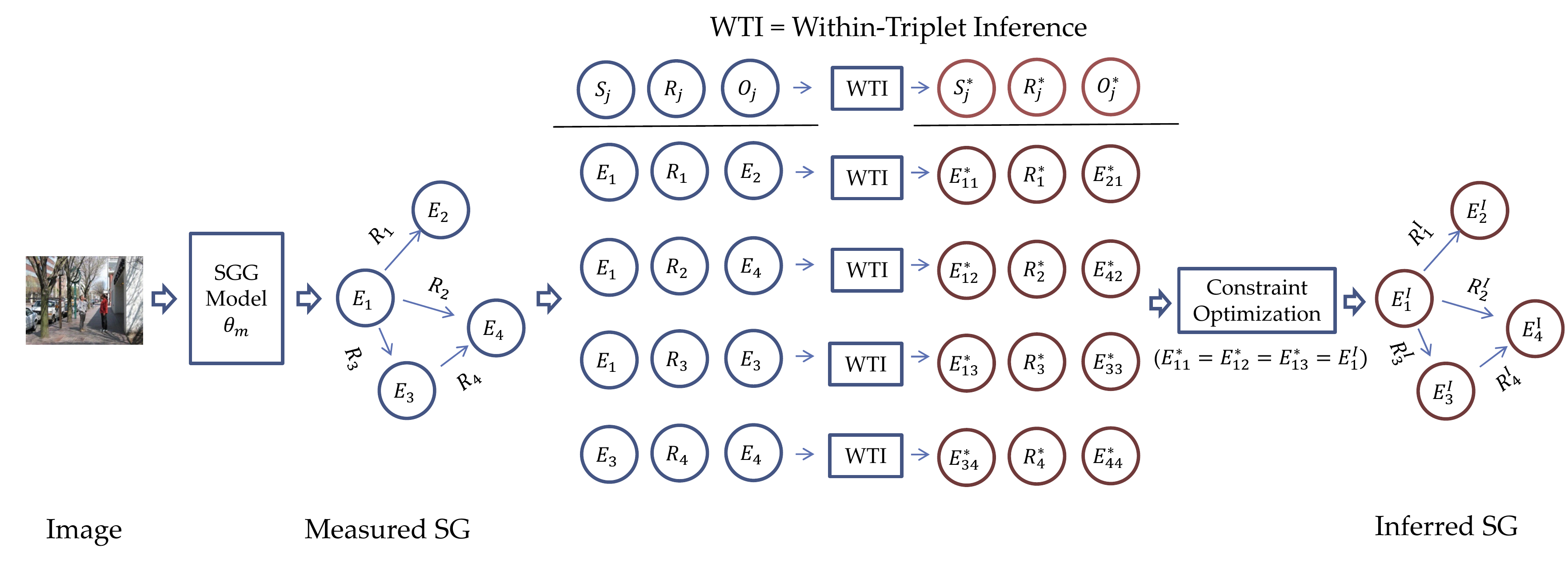}
	\caption{Overview of our proposed approach. For each testing image $I$, SGG baseline model $\theta_m$ generates physically connected triplets with associated uncertainties for the subject, object, and relationship. Our within-triplet inference framework takes all the uncertainties as uncertain evidence and performs posterior inference to infer each triplet separately. Afterward, a constrained optimization procedure is performed to resolve the conflict between object entities.}
	\label{figs:over}
\end{figure*}

The posterior inference is the most efficient probabilistic tool to combine domain-dependent prior with instance-dependent evidence and, to the best of our knowledge, no prior work in SGG literature formulates the problem of triplet generation as a posterior inference problem. The overview of our approach is illustrated in Figure \ref{figs:over}. In summary, our contribution is proposing a posterior inference-based post-processing method where we   
\begin{itemize}
    \item integrate the within-triplet priors with the evidence uncertainties generated by the measurement model and, 
    \item introduce a simple yet novel learning scheme of the within-triplet network where we borrow samples from the semantically similar yet invalid triplet categories. 
\end{itemize}

\section{Related Works}

Our proposed SGG model combines prior knowledge of triplets with the uncertain evidence of measurement models to address the long-tailed issue of relationships. Moreover, we learn the prior model using the similarity of triplets in the language embedding space. Hence, we divide the related works into four major categories as  following

\textbf{Implicit prior and context incorporation}: Global context of an image has been captured either by BiLSTM, Graph Neural Network (GNN), attentional Graph Convolution Network (aGCN), or Conditional Random Field (CRF) in \cite{motif,woo2018linknet,suhail2021energy,xu2021joint,yang2018graph}. Statistical information of triplets is encoded in \cite{xu2017scene,chen2019knowledge,dai2017detecting}. Tang \textit {et al.} \cite{tang2019learning} composed a dynamic tree structure to capture triplet context. Transformer-based context capturing is adopted in \cite{lin2020gps,zareian2020learning,sharifzadeh2020classification,Li_2022_CVPR_sgtr}. Within-triplet relations are incorporated in learnable modules as embeddings in \cite{zhang2017visual,wan2018representation,schroeder2019triplet}. All of these above methods extract the relational information from the training data whereas several other approaches \cite{gu2019scene, yu2017visual, lu2016visual, liang2017deep, zareian2020bridging} rely on external knowledge-base such as ConceptNet \cite{liu2004conceptnet}. 

\textbf{Language models and ambiguity:}  Language and vision modules are combined together to guide the training process in \cite{lu2016visual,yu2017visual,hung2020contextual,peyre2019detecting}. Embeddings from phrasal context \cite{ye2021linguistic},  word embedding-based external knowledge incorporation \cite{gu2019scene,zareian2020bridging}, and caption database \cite{yao2021visual} are used for  supervision. The ambiguity of scene graph triplets is addressed in
\cite{yang2021probabilistic,zhou2020exploring,Li_2022_CVPR_devil}.

\textbf{Long-tail distribution}:  An unbiased metric (\textit{mean recall}) is proposed by \cite{chen2019knowledge,tang2019learning} to measure the performance of the SGG models in \textit{tail} classes. Several works addressed this long-tailedness in training through modified loss functions   \cite{lin2020gps,zareian2020bridging} and re-weighting/re-sampling training samples \cite{desai2021learning, guo2021general,li2021bipartite,Li_2022_CVPR_ppdl,Li_2022_CVPR_devil,chiou2021recovering}.  The most relevant to our work is the inference-time causal intervention proposed by Tang \textit{et al.} \cite{tang2020unbiased} where they identified the `bad bias' from counterfactual causality and attained significantly higher mean recall than other SOTA models. We, instead of applying causal intervention, resort to a graphical model-based approach that can remove the `bad bias' through uncertain evidence insertion into a Bayesian network while maintaining the `good bias' by  within-triplet prior incorporation.  

\textbf{Uncertain evidence:} Incorporation of evidence uncertainty into belief networks has been discussed by Judea Pearl in \cite{pearl2014probabilistic}. A thorough discussion for interested readers can be found in \cite{ chan2005revision}.


\section{Problem formulation}

In scene graph generation database, every image $I$ has an annotation of a scene graph $\mathcal{G}_{I} = (\mathcal{E},\mathcal{R})$ where $\mathcal{E} = \{E_i,B_i\}_{i=1}^{N_E}$ contains the object classes $E_i$ and their bounding boxes $B_i$ whereas $\mathcal{R} = \{R_j(S_j,O_j)\}_{j=1}^{N_T}$ contains the relationships of a scene graph. Each relationship $R_j$ exists between its subject $S_j$ and object $O_j$ where $S_j,O_j \in \{E_i\}$. Now, in training any SGG model parameterized by $\theta$, we can write the cost function $\mathcal{J}(\theta)$ as  
\begin{equation}
    \mathcal{J}(\theta) = \mathbb{E}_{\sim p(I,\mathcal{G})} L(I,\mathcal{G},\theta) \approx  \frac{1}{M}\sum_{i=1}^M  L(I_i,\mathcal{G}_i,\theta)  
\end{equation}
where $\mathbb{E}(\cdot)$ is the expectation operator, $L$ is the loss function with parameter $\theta$ for a sample image $I_i$ with associated scene graph $\mathcal{G}_i$ and $M$ is total number of images. Now, as shown in Figure \ref{figs:long_tailed}, the distribution of scene graphs is skewed towards certain few categories of object pairs and their relationships. Therefore, while training, the cost function $\mathcal{J}(\theta)$ is driven by the \textit{head} pairs, and the relationships dominating in the \textit{tail} pairs are ignored. As a result, many object-conditional distributions of relationships are not preserved after the training is converged. We propose a test-time post-processing method where the object-conditional distribution of relationships is restored by a within-triplet prior Bayesian network. 

\section{Posterior inference of Scene Graphs}

\subsection{Within-triplet Bayesian network}\label{sec:wtb}

\begin{figure}[ht]
	\centering
	\includegraphics[width  =0.5\linewidth]{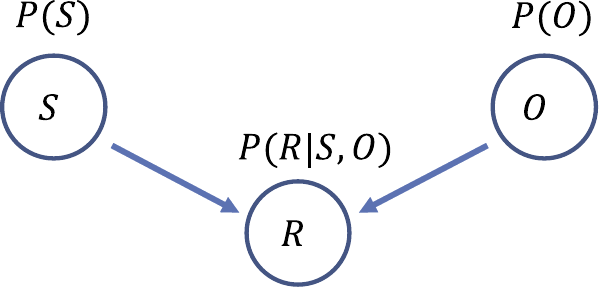}
	\caption{Within-triplet Bayesian network where prior probabilities of subject and object are combined with conditional probability of relationship.}
	\label{figs:WTI}
\end{figure}

A scene graph is a collection of connected triplets and each triplet has three semantic components - subject ($S$), object ($O$), and their relationship ($R$) and two spatial components -  subject bounding box $B_s$ and object bounding box $B_o$. We consider the semantic components as random categorical variables and we aim to model their joint distribution $P(S,R,O)$. We assume there exists no latent con-founder between the subject and object node whereas the relationship node depends on its parent subject and object. Formally, we assume the following statements hold true for any triplet - 
\begin{enumerate}
    \item  Relationship label of a triplet depends on its subject and object $\implies S \rightarrow R \leftarrow O$;
    \item Subject and object   are independent, not given the relationship $\implies S \perp \!\!\! \perp O \not | R$;
    \item Subject and object  becomes dependent, given the relationship $\implies S \not \perp \!\!\! \perp O \mid R$
\end{enumerate} 
Based on these assumptions, we can build a Bayesian network of triplet, as shown in Figure \ref{figs:WTI}, which encodes the joint distribution using the chain rule as follows
\begin{equation}
    P(S,R,O) = P(S)P(O)P(R|S,O) 
\end{equation}
where $P(S)$ and $P(O)$ represent the marginal distribution of the parents $S$ and $O$, and $P(R|S,O)$ represents the conditional distribution of the relationship $R$ given its parent subject and object. This Bayesian network  encodes the prior joint distribution which resides within a triplet and hence we term it as within-triplet Bayesian network. In the next subsection, we discuss how we can debias the measurement probability of a relationship by incorporating them as uncertain evidence into this Bayesian network to perform posterior inference. 

\subsection{Uncertain evidence}\label{sec:uncer_evi}
\begin{figure}[ht]
	\centering
	\includegraphics[width  =0.7\linewidth]{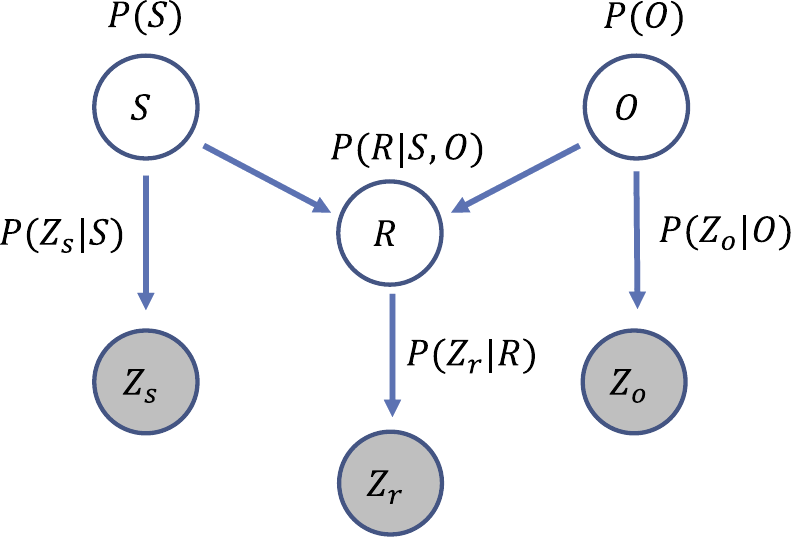}
	\caption{Uncertain evidence of each entity of triplet is incorporated into the Bayesian network as the conditional probability of virtual evidence node denoted as $Z_s, Z_o,$ and $Z_r$.}
	\label{figs:Vert_WTI}
\end{figure}
We denote any trained SGG model with parameter $\theta_m$ which generates measurement probabilities of subject, object, and relationship of every triplet for an image $I$ as $P_{\theta_m,I}(S)$, $P_{\theta_m,I}(O)$, and $P_{\theta_m,I}(R)$. We consider these measurements as uncertain evidence of the nodes in the within-triplet BN in Figure \ref{figs:WTI}. Since this evidence is uncertain, we incorporate it as virtual evidence nodes into the BN as shown in Figure \ref{figs:Vert_WTI}. Following the virtual evidence method proposed by Judea Pearl in \cite{pearl2014probabilistic}, we introduce binary virtual evidence nodes $Z_s,Z_o,$ and $Z_r$ as children of their respective parent evidence nodes $S,R,$ and $O$ and instantiate them as \textit{True}. According to Theorem 5 in \cite{chan2005revision}, the conditional distributions of these virtual nodes maintain the following likelihood ratios 

\begin{equation}
\small
\begin{split}
    &P(Z_s = 1 |s_1 ): .. :P(Z_s = 1 |s_{n} )= \frac{P_{I,\theta_m}(s_1)}{P(s_1)}:..:\frac{P_{I,\theta_m}(s_n)}{P(s_n)}\\ &P(Z_o = 1 |o_1 ): ..  :P(Z_o = 1 |o_n )= \frac{P_{I,\theta_m}(o_1)}{P(o_1)}:..:\frac{P_{I,\theta_m}(o_n)}{P(o_n)}\\ &P(Z_r = 1 |r_1 ): ..  :P(Z_r = 1 |r_n )= \frac{P_{I,\theta_m}(r_1)}{P(r_1)}:..:\frac{P_{I,\theta_m}(r_n)}{P(r_n)}
\end{split}
\end{equation}
where $P(S=s), P(O=o),$ and $P(R=r)$ are the marginal probabilities of subject, object, and relationship node and $P_{\theta_m,I}(S=s)$, $P_{\theta_m,I}(O=o)$, and $P_{\theta_m,I}(R=r)$ are their observed measurement probabilities from image $I$ with model $\theta_m$. Now, we have a complete Bayesian network in Figure \ref{figs:Vert_WTI} with well-defined marginal and conditional probabilities. A brief discussion on uncertain evidence and its incorporation into the Bayesian network is discussed in Appendix A of the supplementary material.  

\subsection{Within-Triplet Inference (WTI) of  triplets}
After the evidence incorporation as virtual evidence nodes, the posterior joint distribution of triplet nodes becomes
\begin{equation}
\small
\begin{split}
    &P(S,R,O|Z_s = 1, Z_o = 1, Z_r = 1) \\
    &\propto P(Z_s=1|S)P(S) P(Z_o = 1|O)P(O) P(Z_r = 1|R)P(R|S,O)\\
    &\propto P_{I,\theta_m}(S) P_{I,\theta_m}(O) \frac{P_{I,\theta_m}(R)}{P(R)} P(R|S,O)
\end{split}
\end{equation}

The Maximum a-Posterior (MAP) of this posterior joint distribution becomes 
\begin{equation}\label{eqs:MAP}
\small
\begin{split}
        &S^*,R^*,O^* = \arg \max_{S,R,O} P(S,R,O|Z_s=1,Z_o=1,Z_r=1) \\
        &= \arg \max_{S,R,O} P_{I,\theta_m}(S) P_{I,\theta_m}(O) \frac{P_{I,\theta_m}(R)}{P(R)} P(R|S,O)
\end{split}
\end{equation}
In Eqn. (\ref{eqs:MAP}), the within-triplet dependency of relationship is encoded in $P(R|S,O)$ and the measurement probability of relationship $P_{I,\theta_m}(R)$ is debiased by its marginal probability $P(R)$.  The subject and object uncertainties are encoded in $P_{I,\theta_m}(S)$ and $P_{I,\theta_m}(O)$. We include some special cases of the MAP Eqn. (\ref{eqs:MAP}) in Appendix B of supplementary material.  
\subsection{Constraint optimization in inferred triplets}\label{sec:const_sat}

Any object entity $E_i$ can reside in multiple triplets as subject $S$ or object $O$ and after individual triplet inference, their inferred values can be different in different triplets. However, to form a valid scene graph, their values should be the same after the inference. Formally, if object entity $E_i$ resides in $J$ triplets, their inferred values $E^*_{ij}$ must satisfy the following constraint 
\begin{equation}
    E^*_{i1} = E^*_{i2} = ... = E^*_{ij} = ... = E^*_{iJ} = E^I_i
\end{equation}
One of the most straightforward ways to satisfy such constraint would be to take the mode of these inferred values as the final value for $E_i$. However, any object entity $E^*_{i}$ should be consistent with respect to all of its connected triplets, and hence we formulate a two-step optimization algorithm to infer $E^*_{i}$ and $R^*_j$ from their connections. 
\paragraph{Object Updating:} In the first step, we infer each object label $E_i^*$ combining its measurement probability $P_{I,\theta_m}(E_i)$  with the within-triplet probabilities of  its connected triplets. We denote $T_i^S$ and $T_i^O$ as the sets of triplets where $E_i$ acts as subject and object respectively 
\begin{equation}\label{eqs:find_trip}
\begin{split}
    T_i^S = \{t_p : t_p(S) = E_i\} \\
    T_i^O = \{t_q : t_q(O) = E_i\} 
\end{split}
\end{equation}
The updated object probability for object $E_i$ and inference of $E_i$ is derived as 
\begin{equation}\label{eqs:const_opt_simple}
\small
\begin{split}
    f (E_i) = P_{I,\theta_m}(E_i) \bigg(&\sum_{t_p \in T_i^S}  P(R = r_{t_p}^I |S=E_i,O = o_{t_p}^I) \\
    +
    &\sum_{t_q \in T_i^O}  P(R = r_{t_q}^I |S = s_{t_q}^I,O=E_i)\bigg)\\
    E^*_i &= \arg \max f (E_i)
\end{split}
\end{equation}
Intuitively speaking, the updated object probability $f (E_i)$  derived in Eqn. (\ref{eqs:const_opt_simple}) combines the uncertain evidence of an object entity $P_{I,\theta_m}(E_i)$ with the prior probabilities of the within-triplet Bayesian networks of all of its connected triplet.

\paragraph{Relationship Updating:} After the first step is completed for each object entity, the conflicts of object entities are resolved with updated entity values. In the second step, we update the relationship label of each triplet based on the updated subject and object values  

\begin{equation}\label{eqn:infer_rel}
\begin{split}
    R_j^* &= \arg \max_{R} P(R_j|Z_r = 1, S=s_j,O=O_j)\\ 
        &= \arg \max_{R} \frac{P_{I,\theta_m}(R_j)}{P(R_j)}  P(R_j|S=s_j,O=o_j)
\end{split}
\end{equation}
Detailed derivation and pseudo-code are provided in Appendix C of the supplementary material. We denote this as \textit{constraint optimization} in our overview in Figure \ref{figs:over}.

\section{Learning BN with embedding similarity}\label{ssec:learn_within} 
\begin{figure}[ht]
	\centering
	\includegraphics[width  =0.6\columnwidth]{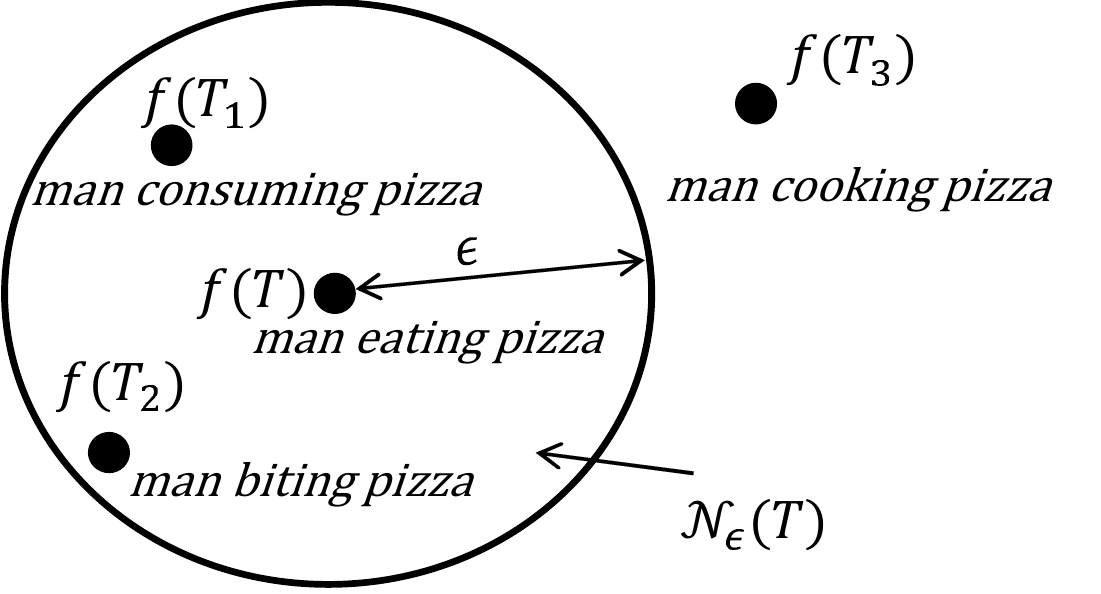}
	\caption{$\mathcal{N}_{\epsilon}(T)$ in the embedding space where $f(T_1),f(T_2) \in \mathcal{N}_{\epsilon}(T)$ and $f(T_3) \notin \mathcal{N}_{\epsilon}(T)$.  }
	\label{figs:emb_neigh}
\end{figure}

The within-triplet priors $P(S) \in \mathbb{R}  ^{N_s}$, $P(O) \in \mathbb{R}^{N_o}$, and conditional distribution $P(R|S,O)  \in \mathbb{R}^{N_s \times N_o \times N_r}$ are learned from annotations of training data where $N_s$, $N_o,$ and $N_r$ denote the number of categories of subject, object, and relationship. A training dataset of $\mathbb{Z}^{3 \times N}$ is created by collecting total $N$ ground truth (GT) triplets from the training images. Afterward, we apply MLE to estimate $P(R|S,O)$ and $P(R)$  as follows
\begin{equation} 
\small
\begin{split}
    P(R=r|S=s,O=o) &= \frac{N^c_{s,r,o}}{\sum_{r^\prime} N^c_{s,r^\prime,o}}\\
    P(R) &= \sum_{S,O} P(R|S,O) P(S)P(O)  
\end{split}
\end{equation}
where $N^c_{s,r,o}$ is the count of triplet with $S=s,O=o$ and $R=r$.  However, because of selecting only top $N_r$ relationships from the training data, many semantically similar triplets whose relationships lie outside these top $N_r$, are ignored in this count (e.g. `man-consuming-pizza' is ignored whereas `man-eating-pizza' is considered as a valid triplet). Hence, we propose a novel sample augmentation method using off-the-shelf sentence embedding models \cite{wolf-etal-2020-transformers} where all the ignored invalid triplets, lying within a $\epsilon$-neighbourhood of a valid triplet in the embedding space, are counted as augmented samples of that valid triplet. For any subject and object pair with $S=s, O=o$, we denote any valid triplet as $T = \{s,r,o\}$ and invalid triplet as $T_i = \{s,r_i,o\}$ where $r \in N_r$ and $r_i \notin N_r$.  Now, if we denote the original count of $T$ as $N^c_{s,r,o}$ and that of $T_i$ as $N_{s,r_i,o}$, we can augment the original count as following 
\begin{equation}\label{eqs:aug_e}
    N^a_{s,r,o} = \begin{cases}
    N^c_{s,r,o} + \sum_{T_i \in \mathcal{N}_{\epsilon}(T)}{N_{s,r_i,o}} \\
    N^c_{s,r,o} \ \ \text{if $\mathcal{N}_{\epsilon}(T) = \emptyset$}
    \end{cases}
\end{equation}
Here, $\epsilon$ is a hyper-parameter and $\mathcal{N}_{\epsilon}(T)$ is defined as the $\epsilon$-neighbourhood of a valid triplet $T$ in the embedding space using the following criteria
\begin{equation}
    \mathcal{N}_{\epsilon}(T) = \{T_i : \phi(f(T),f(T_i)) < \epsilon\}
\end{equation}
where $\phi(f(T),f(T_i))$ represents the distance between two embedding vectors of two triplet $f(T)$ and $f(T_i)$. Since the embeddings lie on a unit-sphere, we employ cosine similarity to measure the angular distance between two embedding vectors of two triplets. We visualize the $\epsilon$-neighbourhood of a valid triplet $T$ in Figure \ref{figs:emb_neigh}. 

\begin{table*}[t]
\small
    
     \begin{tabular}{lllllllll}
     \toprule
     \multirow[]{3}{*}{\textbf{DS}} & \multirow[]{3}{*}{\textbf{Method}} & \multicolumn{6}{c}{Recall and Mean Recall @K} \\
     \cmidrule(r){3-8}
     & &  \multicolumn{2}{c}{PredCls} & \multicolumn{2}{c}{SGCls } & \multicolumn{2}{c}{SGDet} \\  
     \cmidrule(r){3-8}
      & & R@50/100 & mR@50/100 & R@50/100  & mR@50/100 & R@50/100 & mR@50/100 \\
     \midrule
     \multirow{12}{*}{VG} & IMP$^{\diamond}$ \cite{xu2017scene} &  61.63/ 63.63  &  11.53/ 12.38 & 	36.14/ 37.09 &  5.69/ 5.98 &  28.04/ 31.30 &  4.89/ 5.84 \\
       & Inf-IMP & 	59.93/ 62.02 $\downarrow$ & 25.14/ 28.34 $\uparrow$   &  36.02/ 37.09 $\downarrow$ &  12.57/ 14.06 $\uparrow$ &  26.50/ 29.51 $\downarrow$ & 8.58/ 10.67 $\uparrow$ \\
    \cmidrule(r){2-8}
     & MOTIF$^{\diamond}$ \cite{motif} & 59.57/ 63.95 &  12.88/ 15.47 &  36.45/ 38.47 & 7.66/ 8.83 &  26.85/ 30.50 &  5.61/ 6.73 \\
      & Inf-MOTIF &  51.49/ 55.07 $\downarrow$ &  24.67/ 30.71 $\uparrow$ & 32.15/ 33.79 $\downarrow$ &  14.50/ 17.40 $\uparrow$ &  23.94/ 27.09 $\downarrow$  &  9.36/ 11.71 $\uparrow$\\
    \cmidrule(r){2-8}
       & VCTree$^{\diamond}$  \cite{tang2019learning} &  65.46/ 67.18  & 15.36/ 16.61 &  44.15/ 45.11 & 9.17/ 9.83 &  29.94/ 32.57 & 6.21/ 6.96 \\
      & Inf-VCTree &  59.50/ 60.97 $\downarrow$ &  28.14/ 30.72 $\uparrow$ &  40.69/ 41.55 $\downarrow$ &  17.31/ 19.40 $\uparrow$ &  27.74/ 30.10 $\downarrow$ & 10.40/ 11.86 $\uparrow$\\
    \cmidrule(r){2-8}
       & Unb-MOTIF$^{\diamond}$ \cite{tang2020unbiased} &  45.87/ 51.24 &  24.75/ 28.69 &  26.30/ 28.78 &  13.21/ 15.06 & 16.25/ 19.53 &  8.65/ 10.47 \\
     
      & Inf-Unb-MOTIF &  42.40/ 46.82 $\downarrow$ &  28.64/ 35.65 $\uparrow$ &  24.13/ 26.28 $\downarrow$ &  15.85/ 18.88 $\uparrow$ &  15.06/ 18.03 $\downarrow$ &  9.60/ 11.94 $\uparrow$\\
      \cmidrule(r){2-8}
       & DLFE-MOTIF$^{\blacklozenge}$ \cite{chiou2021recovering} &  51.63/ 53.28  &  26.87/ 28.75 &  28.79/ 29.66 &  15.61 /16.38 &  24.22/ 27.95 & 10.62/ 12.61 \\
      & Inf-DLFE-MOTIF &  43.27/ 44.82 $\downarrow$ &  35.25/ 38.20 $\uparrow$ &  24.34/ 25.14 $\downarrow$
 &  19.74/ 20.66 $\uparrow$ & 20.61/ 23.80 $\downarrow$ & 14.07/ 16.76 $\uparrow$\\
      \cmidrule(r){2-8}
      & BGNN$^{\blacklozenge}$  \cite{li2021bipartite} &  58.15/ 60.41  & 29.46/ 31.83  &  -/ - &  - /- &  30.26/ 34.98 &  10.37/ 12.31 \\
      & Inf-BGNN  &  55.42/ 57.47 $\downarrow$  & 32.18/ 34.27 $\uparrow$ &  -/ - &  - /- &  26.16/ 30.11 $\downarrow$ &  13.24/ 16.10 $\uparrow$ \\
      \midrule
      
       \multirow{8}{*}{GQA} & IMP$^{\diamond}$ \cite{xu2017scene} &  61.94/ 63.68 & 13.04/ 13.74 &  34.25/ 34.83 &  7.46/ 7.80 & 25.39/ 27.42 & 5.77/ 6.58 \\
      & Inf-IMP &  61.87/ 63.98 $\uparrow$ &  35.14/ 37.54 $\uparrow$ &  33.19/ 34.01 $\downarrow$&  19.06/ 20.33 $\uparrow$ &  23.46/ 25.56 $\downarrow$ & 12.17/ 14.13 $\uparrow$ \\
     \cmidrule(r){2-8}
      & MOTIF$^{\diamond}$ \cite{motif} & 68.29/ 69.65 &  20.67/ 21.56 &  34.89/ 35.43 & 10.90/  11.31 &  27.83/ 29.38 &  7.38/ 8.32 \\
      & Inf-MOTIF   &  62.95/ 64.23 $\downarrow$ &  37.93/ 40.07 $\uparrow$ &  31.82/ 32.35 $\downarrow$ &  19.09/ 20.00 $\uparrow$ &  25.51/ 26.86 $\downarrow$ &  14.34/ 15.84 $\uparrow$\\
    \cmidrule(r){2-8}
     & VCTree$^{\diamond}$  \cite{tang2019learning} &  68.83/ 70.14 &  22.07/ 23.01 &  35.04/ 35.58 &  10.59/ 10.97 &  27.21/ 28.79 &  7.03/ 7.75\\
      & Inf-VCTree  &  62.80/ 64.05 $\downarrow$ &  39.44/ 41.63 $\uparrow$ & 32.23/ 32.80 $\downarrow$ &  19.18/ 20.03 $\uparrow$ &  25.04/ 26.44 $\downarrow$ &  13.58/ 15.11 $\uparrow$\\
    \cmidrule(r){2-8}
      & Unb-MOTIF$^{\diamond}$ \cite{tang2020unbiased} &  51.87/ 55.87 &  27.81/ 32.30 &  26.10/ 28.06 &  14.09/ 16.33 &  18.22/ 21.63 &  10.78/ 12.89 \\
      & Inf-Unb-MOTIF  &  49.86/ 53.60 $\downarrow$ &  34.45/ 40.80 $\uparrow$ & 24.76/ 26.74 $\downarrow$ &  17.18/ 20.59 $\uparrow$ &  16.84/ 19.94 $\downarrow$ &  12.35/ 14.76 $\uparrow$\\
     \bottomrule
    \end{tabular}
    \caption{\textbf{R@K} and \textbf{mR@K} results of inference with prefix `Inf-'. We  observe a significant increase in \textbf{mR@K} with a slight decrease in \textbf{R@K} for all baseline models in both datasets. Graph constraint is applied in all settings. Baseline results are generated by codebase released by \cite{tang2020unbiased} ($\diamond$) and by respective authors ($\blacklozenge$)}.
    \label{tab:Mean_Recall_Refined}
\end{table*}
\section{Experimental settings}

\subsection{Dataset}\label{sec:dataset}
We evaluate our proposed method on two datasets: (1) Visual Genome (\textbf{VG}),  and (2) \textbf{GQA}.  

\textbf{(1) Visual Genome}:
For SGG, the most commonly used database is the Visual Genome (\textbf{VG}) \cite{krishna2017visual} and from the original database, the most frequent $150$ object and $50$ predicate categories are retained \cite{xu2017scene,motif}. We adopt the standard train-test split ratio of $70:30$. The number of prior triplets from the original training dataset is around $323K$ and after augmenting with $\epsilon = 0.05$, the number rises to around $391K$. 

\textbf{(2) GQA}: GQA \cite{hudson2019gqa} is a refined dataset derived from the VG images. We retain images only with the  most frequent $150$ object and $50$ relationship categories. We train on around $50k$ valid images and perform the evaluation on the validation dataset of valid $7k$ images. We collect over $190K$ prior triplets and after embedding-based augmentation with $\epsilon = 0.05$, the number rises to around $200K$.  

\subsection{Task description}\label{sect:task}
A triplet is considered correct if the subject, relationship, and object label matches with the ground truth labels and the boundary boxes of subject and object have an Intersection over Union (IoU) of at least $50\%$ with the ground truth annotations. We consider three test-time tasks, defined by \cite{xu2017scene}, - (1) \textbf{PredCls:} known object labels and locations, (2) \textbf{SGCls:}  known object locations, and (3) \textbf{SGDet:} where no information about objects are known. We apply graph constraints for all the tasks where for each pair of objects, only one relationship is allowed.

\subsection{Evaluation metrics}
The performance of our debiased SGGs is evaluated through recall (\textbf{R@K}) and mean recall (\textbf{mR@K}). \textbf{R@K} of an image is computed as the fraction of ground truth triplets in top@K predicted triplets \cite{lu2016visual} whereas \textbf{mR@K} computes recall for each relationship separately and then the average over all relationships are computed \cite{chen2019knowledge,tang2019learning}.

\subsection{Implementation details }\label{sec:imp_detail}
We perform training and testing of the baseline models  released by \cite{tang2020unbiased}, \cite{chiou2021recovering}, and \cite{li2021bipartite}. We collect measurement results  and ground truth annotations of the testing database using Python. For sample augmentation, we employ the sentence transformer model `all-mpnet-base-v2' released by HuggingFace \cite{wolf-etal-2020-transformers}. We learn the within-triplet prior from the original and augmented annotations, and perform
posterior inference in MATLAB on a computer with core i5 7th generation Intel processor running at $2.5$ MHz with $8.00$ GB RAM. The total training time for the prior probabilities of \textbf{VG} dataset is $968s$ and that of \textbf{GQA} is $430s$. The  inference task per image requires $0.13s$. 

\section{Experimental results}
\subsection{Quantitative results} 
We generate triplet measurements from four classical SGG models - (1) IMP \cite{xu2017scene}, (2) MOTIF \cite{motif}, (3) VCTree \cite{tang2019learning}, and (4) Unb-MOTIF \cite{tang2020unbiased} from codebase  \cite{tang2020unbiased} ($\diamond$), and two recent-most SOTA SGG models released by (1) DLFE-MOTIF \cite{chiou2021recovering} and (2) BGNN \cite{li2021bipartite} ($\blacklozenge$).  Considering these measurements as uncertain evidence, we perform within-triplet inference for all three settings in Sect. \ref{sect:task} and conduct two-step updating only for \textbf{SGCls} and \textbf{SGDet} settings. We denote the final  results with the prefix `Inf-'. The Bayesian network is learned from the augmented counts of triplet derived by Eqn. (\ref{eqs:aug_e}). We report \textbf{R@K} and \textbf{mR@K} for all three tasks with \textbf{VG} and \textbf{GQA} in  Table \ref{tab:Mean_Recall_Refined}. We also include a separate comparison with other bias removal techniques (1) \textbf{Unb-}  \cite{tang2020unbiased}, (2)  \textbf{DLFE-} \cite{chiou2021recovering}, and (3) \textbf{NICE-\cite{Li_2022_CVPR_devil}} in Table \ref{tab:comp_causal}. Our method performs better in balancing the \textit{head} and \textit{tail} classes without any retraining of the biased model.   

\begin{table}[ht]
\small
\centering
\begin{tabular}{llll}
    \toprule
     \multirow[]{2}{*}{\textbf{Method}}
     & \multirow[]{2}{*}{\textbf{Re-train}} & \multicolumn{1}{c}{R@K} & \multicolumn{1}{c}{mR@K }\\
     \cmidrule(r){3-4}
      & & @50/100 & @50/100 \\
    \midrule
   DT2-ACBS \cite{desai2021learning} & \textbf{Yes} & 23.3/ 25.6 & \textbf{35.9/ 39.7} \\
   \midrule
    BGNN \cite{li2021bipartite}& \textbf{Yes} & \textbf{59.2/ 61.3} & 30.4/ 32.9  \\
    \midrule
    IMP\cite{xu2017scene} & - & 61.6/ 63.6 & 11.5/ 12.4 \\
    Inf-IMP (Ours) & \textbf{No} & 59.9/ 62.0 & 25.1/ 28.3 \\
    \midrule
    MOTIF\cite{xu2017scene} & - & 59.6/ 64.0 & 12.9/ 15.5 \\
    Unb-MOTIF\cite{tang2020unbiased} & \textbf{No} & 45.9/ 51.2 & 24.8/ 28.7 \\
    DLFE-MOTIF\cite{chiou2021recovering} & \textbf{Yes} & 51.6/ 53.2 & 26.9/ 28.8\\
    NICE-MOTIF \cite{Li_2022_CVPR_devil} & \textbf{Yes} &  \textbf{55.1/ 57.2} & \textbf{29.9/ 32.3} \\
    Inf-MOTIF (Ours) & \textbf{No} & 51.5/ 55.1 & 24.7/ 30.7 \\
    \midrule
     VCTree\cite{tang2019learning} & - & 65.5/ 67.2 & 15.4/ 16.6 \\

     Unb-VCTree \cite{tang2020unbiased} & \textbf{No} & 47.2/ 51.6 & 25.4/ 28.7   \\
    DLFE-VCTree \cite{chiou2021recovering} &  \textbf{Yes} &   51.8/ 53.5 &   25.3/ 27.1 \\
    NICE-VCTree \cite{Li_2022_CVPR_devil} & \textbf{Yes} &   55.0/ 56.9  &   \textbf{30.7/ 33.0} \\
     Inf-VCTree (Ours) & \textbf{No} & \textbf{59.5/ 61.0} & 28.1/ 30.7 \\
   \midrule
    \end{tabular}
    \caption{Comparison with other de-biasing methods in PredCls. Without re-training, our loss in \textbf{R@K} is significantly lower, and gain in \textbf{mR@K} is higher or competitive than other SOTA debiasing methods.}\label{tab:comp_causal}
\end{table}

\subsection{Analysis}

\subsubsection{Ablation study on prior and uncertain evidence}
\begin{figure*}[hbtp]
	\centering
	\begin{subfigure}[b]{0.49\textwidth}
	\centering
	\includegraphics[width=1\textwidth]{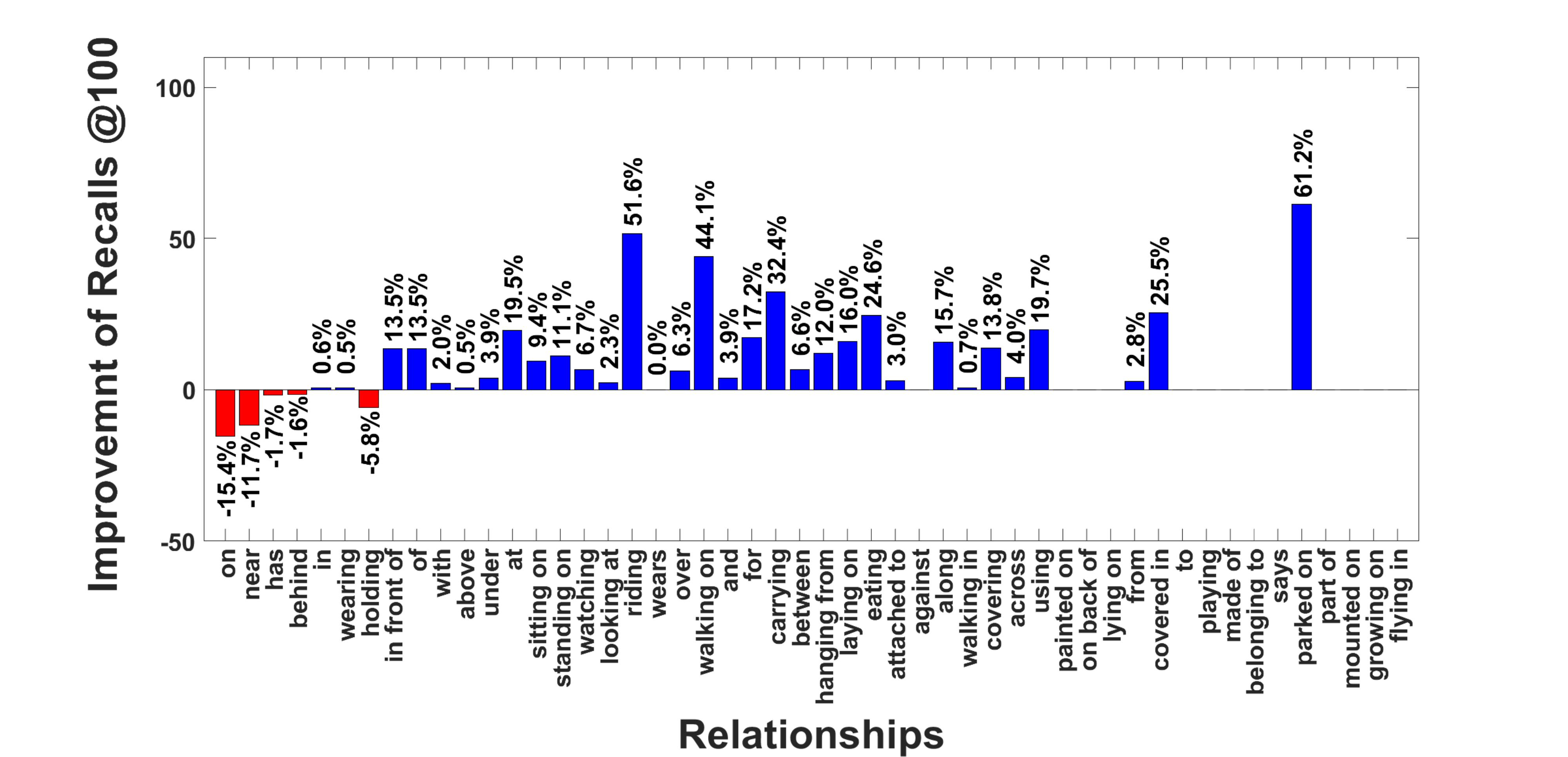}
	\caption{Recall improvement of each relationship with original samples.  }
	\label{figs:improv_base_org}
	\end{subfigure}
	\begin{subfigure}[b]{0.49\textwidth}
		\centering
		\includegraphics[width=1\textwidth]{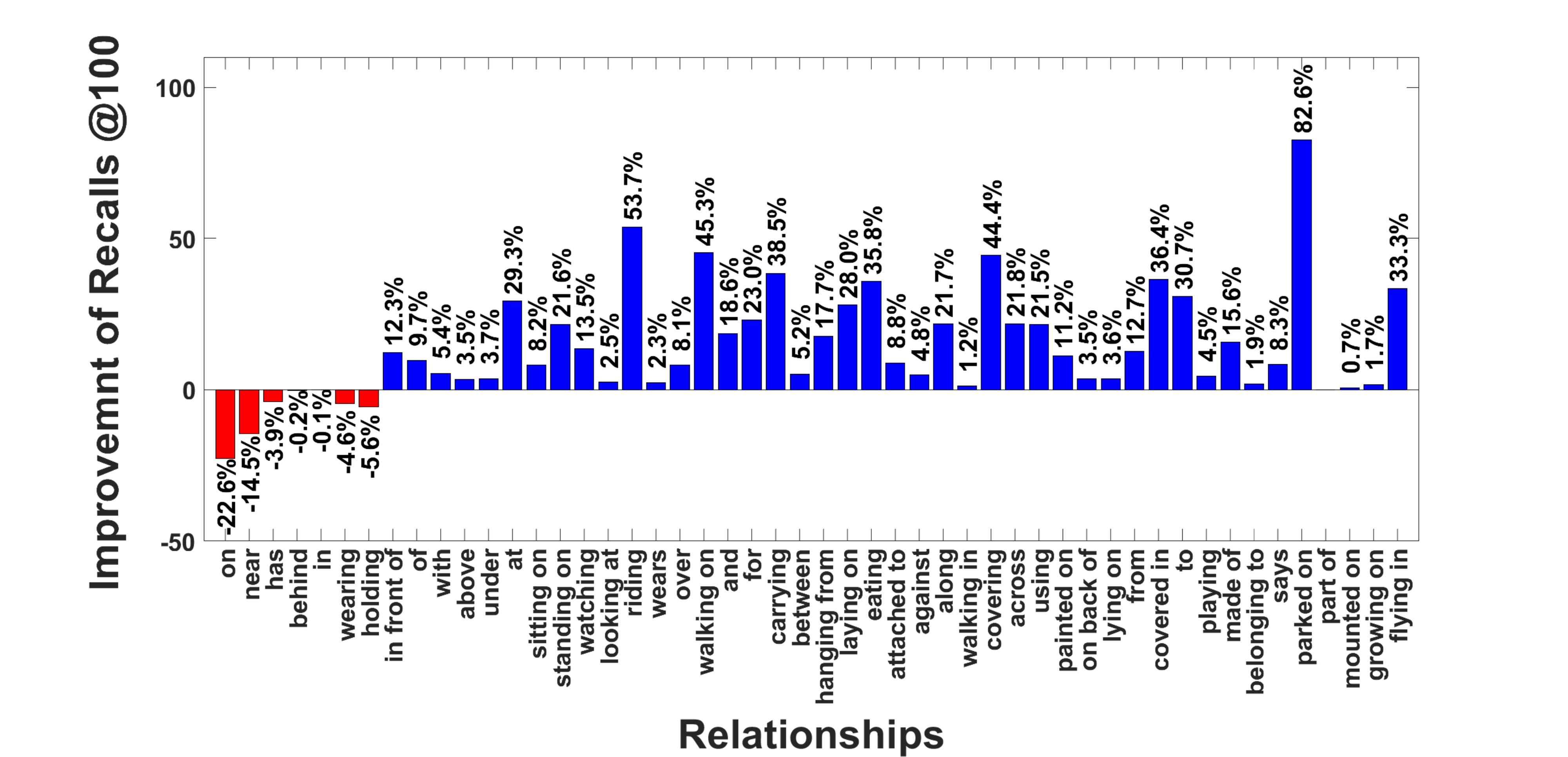}
	\caption{Recall improvement of each relationship with augmented samples. }
    \label{figs:improv_base_aug}
	\end{subfigure}
	\caption{Improvement of mean recalls with VCTree \cite{tang2019learning} evidence for PredCls task in VG. Relationships are ordered in descending order of their frequencies. In (a),  BN learned with original samples improves the \textit{mid} region whereas in (b),  the embedding-based augmentation improves both the \textit{mid} and the \textit{tail} ones. In both cases, the \textit{head} relationships are worsened after debiasing.}
\end{figure*}

We perform an ablation study on the measurement results of the PredCls task by VCTree \cite{tang2019learning} in Table \ref{tab:abla_pri}. We visualize the improvement of tail classes with BN learned from original and augmented samples in Figure \ref{figs:improv_base_org} and \ref{figs:improv_base_aug}. In the former case, \textit{mid} relationships are improving after inference and in the latter case both \textit{mid} and \textit{tail} are improving.

\begin{table}[t]
\small
\centering
\begin{tabular}{ lll}
     \toprule
     \multirow[]{2}{*}{\textbf{Method}}
     & \multicolumn{1}{c}{R@K} & \multicolumn{1}{c}{Mean R@K }\\
     \cmidrule(r){2-3}
      &  @50/100 & @50/100 \\
      \midrule
      Uncertain evidence (VCTree) &  65.5/ 67.2 & 15.4/ 16.6\\
      \midrule
      WT BN only (org) (FREQ) &  64.3/ 65.8 $\downarrow$ &   16.1/ 17.5 $\uparrow$\\
      \midrule
     WT BN only (aug) 
     (Ours) &  62.7/ 64.2 $\downarrow$ &  16.3/ 17.8 $\uparrow$\\
      \midrule
     WT BN (org) +  Unc. Evi (Ours) &  62.5 /64.1 $\downarrow$ &  22.7/ 24.8 $\uparrow$\\
      \midrule
     WT BN (aug) + Unc. Evi (Ours) &  59.5 /61.0 $\downarrow$ &  28.1/ 30.7 $\uparrow$\\
      \bottomrule
    \end{tabular}
    \caption{Ablation study on PredCls performance for VG. We observe the consistent improvement of \textbf{mR@K} starting with uncertain evidence and ending in posterior inference with uncertain evidence with BN learned from augmented samples. }\label{tab:abla_pri}
\end{table}
\subsubsection{Ablation study on conflict resolution}
As discussed in Sect. \ref{sec:const_sat}, the potential conflicts of object labels after within-triplet inference can be resolved with naive mode-selection or by our proposed constraint optimization. We observe the effectiveness of the proposed  optimization method over mode selection in Table \ref{tab:abla_const}.
\begin{table}[ht]
\small
\centering

\begin{tabular}{ lll}
     \toprule
     \multirow[]{2}{*}{\textbf{Method}}
     & \multicolumn{1}{c}{R@K} & \multicolumn{1}{c}{mR@K }\\
     \cmidrule(r){2-3}
      &  @50/100 & @50/100 \\
      \midrule
      VCTree \cite{tang2019learning} & 44.2/ 45.1 &	9.2/ 9.8 \\
      \midrule
      Inf-VCTree (Conflict res. by mode) & 40.3/ 41.2 &	16.9/ 18.7 \\
      \midrule
     Inf-VCTree (Conflict res. by opt.) & 40.7/ 41.6 $\uparrow$ &	17.3/ 19.4 $\uparrow$\\
      \bottomrule
    \end{tabular}
    \caption{Ablation study on conflict resolution for VCTree SGCls performance for Visual Genome. The optimization algorithm performs better than the naive mode-selection version.}\label{tab:abla_const}
\end{table}

\subsubsection{Effect of $\epsilon$ on training data augmentation}

We observe the effect of neighborhood radius $\epsilon$ in the embedding space on \textbf{R@K} and \textbf{mR@K}  in Table \ref{tab:eps_aug}. Larger $\epsilon$ tends to hurt the majority classes more. We choose $\epsilon=0.05$ for our final experiments. 
\begin{table}[ht]
\centering
\small
\begin{tabular}{llllllll}
     \toprule
     \multirow[]{2}{*}{\textbf{Method}}
     & \multirow[]{2}{*}{$\epsilon$} & \multicolumn{3}{c}{Recall@K} & \multicolumn{3}{c}{ Mean Recal@K }\\
     \cmidrule(r){3-8}
      &  &  @50/@100 & @50/@100 \\
    \midrule
     Inf- VCTree (org)  & - & 62.48/ 64.06 & 22.74/ 24.78\\
    \midrule
     \multirow[]{3}{*} {Inf- VCTree (aug)}
    & 0.03  & 59.44/ 60.88 &  28.01/ 30.55  \\
      \cmidrule(r){3-8}
     & 0.05 & 59.50/ 60.97 &  28.14/ 30.72  \\
     \cmidrule(r){3-8}
     & 0.07 & 59.27/ 60.73 &  28.28/ 30.92  \\
     \bottomrule
    \end{tabular}
    \caption{Effect of $\epsilon$ on PredCls performance for VG with VCTree baseline \cite{tang2019learning}. We find that using larger $\epsilon$ tends to drop \textbf{R@K} more. Based on this study, we choose $\epsilon = 0.05$ for our final experiments. }\label{tab:eps_aug}
\end{table}

\subsubsection{Effect on zero-shot recall }
The zero-shot prediction of the measurement model is compromised after posterior inference due to the MLE-based learning of BN. However, since unseen triplets will have higher entropy than the seen ones in the prediction phase, we can filter out the high-entropy triplets so that they do not get refined by the BN. An ablation study with respect to entropy threshold is shown in Table \ref{tab:abla_ent}. 

\begin{table}[ht]
\small
\centering

\begin{tabular}{llll}
     \toprule
     \multirow[]{2}{*}{\textbf{Method}}
     & \multirow[]{2}{*}{\textbf{Entropy Th.}} & \multicolumn{1}{c}{ZS R@100} & \multicolumn{1}{c}{ ZS mR@100 }\\
     \cmidrule(r){3-4}
      & & org/aug & org/aug \\
      \midrule
      VCTree \cite{tang2019learning} & - & 6.02 &  16.61 \\
      \midrule
      \multirow[]{4}{*} {Inf-VCTree} & 0 & 6.02/ 6.02 
 &  16.61/ 16.61 \\
 & 1.5 & 5.81/ 6.13 & 20.20/ 22.71\\
 & 2.5 & 5.47/ 6.03 & 24.42/ 29.28 \\
 & 3.912 & 1.06/ 2.30 & 24.78/ 30.70\\
      \bottomrule
    \end{tabular}
    \caption{Zero-shot recall  on PredCls for VG for VCTree \cite{tang2019learning}. Lower thresholds restore the zero-shot capability of measurement models by hurting the mean recall. We choose not to use any threshold (last row) to maximize performance in the mean recall. }\label{tab:abla_ent}
\end{table}

\section{Limitations}
While improving \textbf{mR@K} by our proposed method, we lose performance in \textbf{R@K}. This phenomenon is prevailing in SGG debiasing works since we are perturbing the `head' classes  to gain improvement in the `tail'. Moreover, the augmentation hyper-parameter $\epsilon$ may vary from dataset to dataset and need to be chosen carefully. Another weakness is the compromise in the zero-shot prediction capability of a measurement model due to MLE-based learning of BN. 

\section{Conclusion}
We proposed a debiasing strategy of scene graphs by combining  prior and uncertain evidence of triplets in the Bayesian framework. We performed MAP inference and optimally solved the conflict between object entities to predict the debiased triplet from biased evidence. We augmented the count of valid triplets with semantically similar invalid triplets to alleviate sample insufficiency. Our method showcased significant improvement in mean recall with baseline  measurements. We also attained a better balance between majority and minority performances of the relationships. In the future, we will extend the MAP inference for multiply connected triplets and we will explore well-defined criteria for zero-shot refinement to restore the zero-shot recall of the measurement models. 

{\small
\bibliographystyle{ieee_fullname}
\bibliography{sgg}
}

\end{document}